\pdfoutput=1

\documentclass[11pt]{article}

\usepackage{acl}

\usepackage{times}
\usepackage{latexsym}

\usepackage[T1]{fontenc}

\usepackage[utf8]{inputenc}

\usepackage{microtype}

\usepackage[ruled,linesnumbered]{algorithm2e}
\usepackage{graphicx}
\usepackage{subfigure}
\usepackage{booktabs} 
\usepackage{multirow}
\usepackage{tabularx}
\usepackage{amsmath}

\usepackage{amsmath,amsfonts,bm}

\def\eqref#1{equation~\ref{#1}}

\def\1{\bm{1}}

\def\rvb{{\mathbf{b}}}

\def\rve{{\mathbf{e}}}

\def\rvg{{\mathbf{g}}}
\def\rvh{{\mathbf{h}}}

\def\rvp{{\mathbf{p}}}

\def\rvr{{\mathbf{r}}}

\def\rmE{{\mathbf{E}}}

\def\rmN{{\mathbf{N}}}
\def\rmO{{\mathbf{O}}}

\def\rmW{{\mathbf{W}}}

\DeclareMathAlphabet{\mathsfit}{\encodingdefault}{\sfdefault}{m}{sl}
\SetMathAlphabet{\mathsfit}{bold}{\encodingdefault}{\sfdefault}{bx}{n}

\def\gE{{\mathcal{E}}}

\def\gG{{\mathcal{G}}}

\def\gR{{\mathcal{R}}}
\def\gS{{\mathcal{S}}}
\def\gT{{\mathcal{T}}}

\title{Deep Reinforcement Learning for Entity Alignment}

\author{Lingbing Guo$^{1,2,3}$ \and Yuqiang Han$^{1,2}$ \and Qiang Zhang$^{1,2}$\thanks{$^*$Corresponding authors.} \and Huajun Chen$^{1,2,3*}$ \\
\textsuperscript{1} College of Computer Science and Technology, Zhejiang University \\  \textsuperscript{2} Hangzhou Innovation Center, Zhejiang University\\ \textsuperscript{3} Alibaba-Zhejiang University Joint Reseach Institute of Frontier Technologies \\ \texttt{\{lbguo, hyq2015, qiang.zhang.cs, huajunsir\}@zju.edu.cn} }

\begin{document}
\maketitle
\begin{abstract}
Embedding-based methods have attracted increasing attention in recent entity alignment (EA) studies. Although great promise they can offer, there are still several limitations. The most notable is that they identify the aligned entities based on cosine similarity, ignoring the semantics underlying the embeddings themselves. Furthermore, these methods are shortsighted, heuristically selecting the closest entity as the target and allowing multiple entities to match the same candidate. To address these limitations, we model entity alignment as a sequential decision-making task, in which an agent sequentially decides whether two entities are matched or mismatched based on their representation vectors. The proposed reinforcement learning (RL)-based entity alignment framework can be flexibly adapted to most embedding-based EA methods. The experimental results demonstrate that it consistently advances the performance of several state-of-the-art methods, with a maximum improvement of 31.1\% on Hits@1.
\end{abstract}

\section{Introduction}
\label{sec:intro}
Entity alignment (EA) is one of the most crucial tasks in knowledge graph (KG) studies. It aims to seek the potentially aligned entity pairs between two KGs, such that distributed knowledge can be linked for better supporting downstream applications. Generally, a fact in a KG can be represented by a triplet $(e_x^1, r^1, e_y^1)$, where $e_x^1$, $e_y^1$ denote the head and tail entities in the first KG $\gG_1$. $r^1$ is the relation connecting them. With a small number of known alignment pairs as anchors, embedding-based entity alignment (EEA) methods can learn the representations of entities belonging to respective KGs in a unified space and exploit the underlying aligned pairs based on the embedding distance. For example, $e_x^2$ will be chosen as target entity for $e_x^1$ if its embedding is closest to the embedding of $e_x^1$ in vector space.

\begin{figure}[t]
	\centering
	\includegraphics[width=\linewidth]{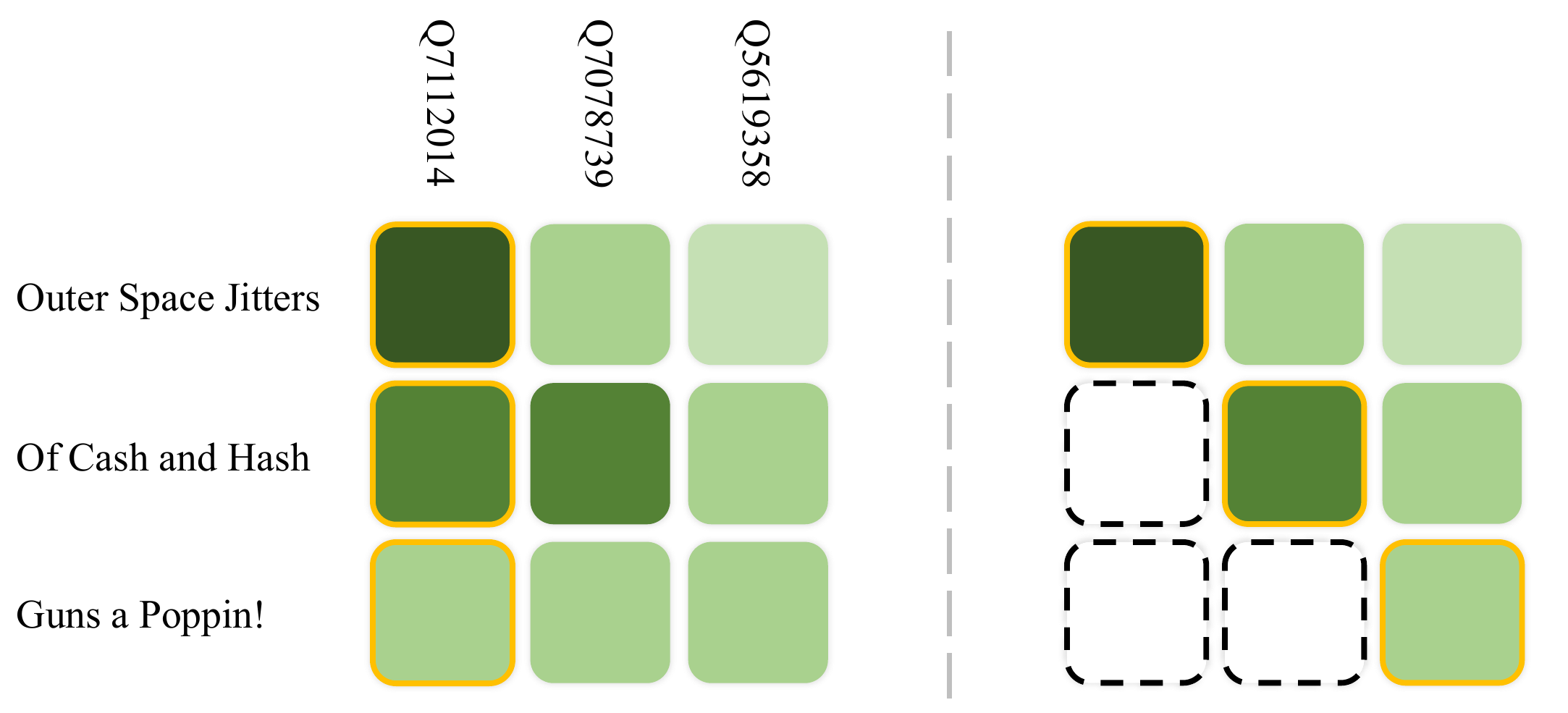}
	\caption{Different evaluation strategies. The ranking strategy (left) heuristically selects the candidate with the largest similarity. The sequential strategy (right) allows each candidate to be matched only once. Deeper color indicates higher similarity. Diagonals are correct matches. Cells with yellow borders are the selected entities, while those with dotted borders denote the excluded entities.}
	\label{fig:ea_example}
	\vspace{-1.2em}
\end{figure}

Although recent EEA methods~\cite{MTransE,JAPE,BootEA,RSN,RDGCN,BertINT,EntityPairEmbedding,AliNet} have made great performance improvement, they rarely consider the evaluation process. For example, in Figure~\ref{fig:ea_example}, all three films are directed by Jules White and have similar casts. This makes the EEA methods confused to discriminate the true aligned entities from other candidates. Current ranking strategy heuristically chooses the nearest entities without considering that some entities have already been matched before. An entity with the largest similarity is not always the true target, especially when this candidate has been matched with other entities. In contrast, we can model entity alignment as a sequential decision-making task, where the agent sequentially decides whether a candidate embedding is aligned with the input one. Then, the environment will exclude the matched candidates in the subsequent decisions.

One issue with the sequential strategy is the accumulated errors. Due to the heterogeneity of KGs, a pair of underlying aligned entities may not share an identical neighborhood. This makes their embeddings not as similar as desired with each other (e.g., ``Guns a Poppin!'' in Figure~\ref{fig:ea_example}). But semantics in the embeddings may still indicate the actual target. It is worth estimating the alignment score directly from their embeddings. It is also important to negate a most likely candidate for maximizing the long-term rewards.

In this paper, we draw on the insights of reinforcement learning (RL) that has recently received great attention in many fields~\cite{DQN,DDPG,AlphaGo}. 
With the trained embeddings of any existing EEA models as raw input, we train an agent to find as many alignment pairs as possible to maximize the reward. Meanwhile, we adopt a curriculum learning~\cite{CurriculumLearning} strategy for the environment to provide candidate entity pairs as observations of increasing difficulty. In sum, our contributions are three-fold:

\begin{itemize}
	\item We propose to model entity alignment as a sequential decision-making task. To the best of our knowledge, this is the first method that provides a general solution to improve the evaluation strategy for the EEA task.
	\item We implement an end-to-end RL-based entity alignment (RLEA) framework to solve the sequential EEA problem\footnote{https://github.com/guolingbing/RLEA}. We elaborate an entity alignment environment to sample candidate pairs as observations efficiently. Besides, we design a policy network that takes self-embedding, neighborhood, and long-term rewards into account.
	\item We conduct extensive experiments to show that RLEA can significantly and consistently improve the state-of-the-art EEA methods.
\end{itemize}

\section{Related Work}
\label{sec:background}

\subsection{Embedding-based Entity Alignment}
We divide the exiting EEA methods into two categories. The first is based on the well-known KG embedding method TransE~\cite{TransE}. TransE models a triplet $(e_x^1, r^1, e_y^1)$ as $\rve_x^1 + \rvr^1 \approx \rve_y^1$, with the boldfaced as the corresponding embeddings. Many methods use TransE as the KG embedding model for the EA task: MTransE~\cite{MTransE} sets a learnable matrix to project the entity embeddings from the source KG to the space of the target KG. Then, the distance among entity embeddings from different KGs can be used to estimate the similarity. This idea is extended by later works, e.g., KDCoE~\cite{KDCoE}, SEA~\cite{SEA}, and OTEA~\cite{OTEA}. Specifically, KDCoE learns the triplet embedding model and the description embedding model in a co-training fashion. SEA leverages adversarial learning to learn better projection matrix. It also considers the attribute information. OTEA makes use of optimal transport theories to advance the learning process of MTransE. On the other hand, JAPE~\cite{JAPE} and IPransE~\cite{IPTransE} adopt a mapping strategy that utterly different from MTransE. They directly set two entities in a known alignment pair to one embedding vector. Therefore, the vector spaces of two KGs are naturally connected. For example, given a known alignment $(e^1, e^2)$, $e^1$, $e^2$ will be mapped to one embedding vector $\mathbf{e}$. 

The other line of EEA research focuses on the design of embedding models. Great efforts were put into graph convolutional networks (GCNs)~\cite{GCNs}, e.g.,  GCN-Align~\cite{GCN-Align}, RDGCN~\cite{RDGCN}, and graph attention networks (GATs)~\cite{GAT}, e.g., MuGNN\cite{MuGNN}, AliNet~\cite{AliNet}. Most of them adopt the mapping strategy to map entities in each known pair to one vector to connect two KGs. Therefore, these methods center on the design of different graph network structures, which is out of the discussion of this paper. We refer the readers to~\cite{OpenEA,Survey} for details.

One unique method, BootEA~\cite{BootEA}, iteratively labels likely entity alignment as training data. BootEA is a powerful method that greatly improved the performance of the basic AlignE model. This bootstrapping method is closely related to RLEA as it also assumes that a candidate entity should not be matched more than once. However, BootEA does not have a learning process, similar to its followers~\cite{CoordinatedReasoning,CollectiveEA,RAGA}. The entity alignment pairs are computed based on the cosine similarity and further threshed by a hyper-parameter to filter out those with low similarity. On the other hand, the bootstrapping algorithm must run with the embedding model iteratively, making BootEA more sensitive to parameter settings. Nevertheless, there is no contradiction in integrating RLEA with BootEA to achieve better performance (see 
Section~\ref{app:bootea}).

The above methods have different objectives and investigate diverse techniques. However, RLEA only needs their trained embeddings as input data, which is sufficient to achieve much better performance on several datasets.

\subsection{Deep Reinforcement Learning for Knowledge Graphs}
One most relevant work to this paper is CEAFF~\cite{CEAFF}, which also leverages RL algorithms and believes in 1-to-1 alignment. But CEAFF focuses more on generating and integrating different entity features. The RL part is less explored. From its experimental results~\cite{CEAFF}, we can find that RL-based CEAFF only outperformed its heuristic version slightly. Moreover, CEAFF does not provide a general solution for sequential EEA task. It is not applicable to most existing EEA methods.

DeepPath~\cite{deeppath} and its followers~\cite{RL-reasoning2,RL-reasoning1} are also well-known RL-based KG embedding methods. They leverage RL agents to continually extend paths for multi-hop reasoning. There are two major differences. First, DeepPath focuses more on the design of the reward function. It takes accuracy, diversity, and efficiency into consideration when estimating an action's reward. However, the design of the environment is relatively straightforward, as the next state is certain after receiving an edge as action. By contrast, the reward in our sequential EA task can be simply assigned by comparing the output action (i.e., \emph{match} or \emph{mismatch}) with the actual label; nevertheless, any valid entity pair can be set as next state. Therefore, we focus more on formalizing this problem and building a proper environment where the agent can explore efficiently.

Additionally, some methods like KAGAN~\cite{KAGAN} only use the policy gradient algorithm~\cite{REINFORCE} to update their network parameters. They do not really learn a policy to solve a sequential decision-making problem. Therefore, we do not review them in this paper.

\begin{figure*}[t]
	\centering
	\includegraphics[width=\linewidth]{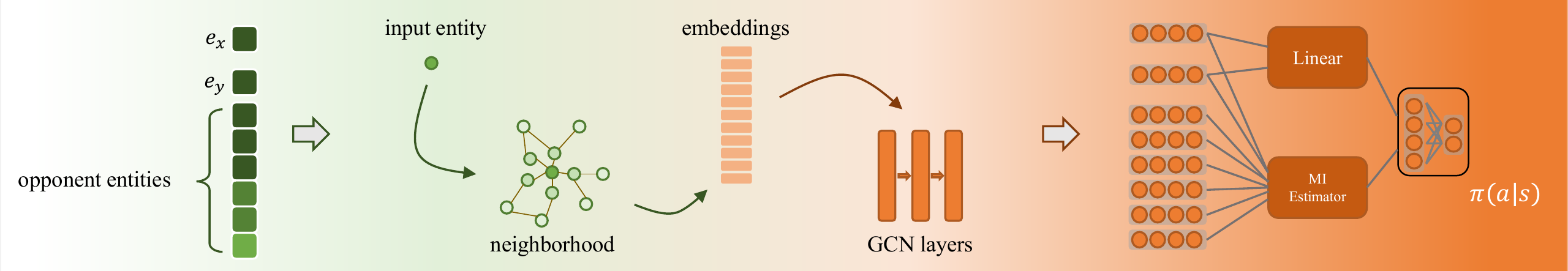}
	\caption{Overview of the policy network. We first use a GNN model to aggregate the neighbor embeddings of each entity. The output representations of $e_x$, $e_y$ are then fed into a linear layer that maps features to an unnormalized estimation of the alignment score. We also leverage a mutual information estimator, which takes the opponent entity representations as negative examples. We combine the output of two types of estimations to obtain the final action distribution.}
	\label{fig:agent}
	\vspace{-1.2em}
\end{figure*}

\section{Methodology}
\label{sec:methodology}

\subsection{Preliminaries}
Let $\gG_1 = \{\gE_1, \gR_1, \gT_1\}$ and $\gG_2 = \{\gE_2, \gR_2, \gT_2\}$ be the source and target KGs, with $\gE$, $\gR$, $\gT$ denoting the entity, relation, and triplet sets respectively. The proposed RL-based framework consists of two modules, i.e., the agent and the environment. We use the trained entity embeddings $\rmE_1$, $\rmE_2$ of any EEA models as input for the agent. The training set, same to the existing works, is still a small number of known entity alignment $\gS \subset \gE_1 \times \gE_2$ provided by the dataset. 

In each training episode, the states and actions are generated by the environment and the agent in an alternative order, i.e., $s_1, a_1, s_2, a_2, ..., s_i, a_i$. We define a state $s$ as a pair of arbitrary entities $[e_x, e_y]$ belonging to respective KGs (we rewrite $[e_x^1, e_y^2]$ as $[e_x, e_y]$ for readability, the same below). An action $a \in \{0, 1\}$ represents the decision of the agent that indicates \emph{match} or \emph{mismatch} for $[e_x, e_y]$. Each state also has a label $l \in \{0, 1\}$, implying the right decision. It is worth noting that an action may still have a positive effect even if it is not equal to the label. For example, an incorrect \emph{match} action can also exclude two wrong entities correctly. 

In the following sections, we call the case of $a=0 \wedge l=0$ a \emph{true mismatch}, $a=0 \wedge l=1$ a \emph{false mismatch}, $a=1 \wedge l=0$ a \emph{false match}, and $a=1 \wedge l=1$ a \emph{true match}. Therefore, the number of correct aligned entity pairs equals to that of \emph{true match}, which is proportional to the Hits@1 result in the conventional EEA task.

\subsection{Agent}

We start by introducing the agent module, which is modeled by neural networks.

\paragraph{State}
A state $s = [e_x, e_y]$ is given by the environment. We take the following features into consideration: (1) the embeddings of two entities $\rve_x$, $\rve_y$; (2) the neighbor embedding sets of two entities $\rmN_x$, $\rmN_y$; (3) the opponent entity embedding set $\rmO_y$ of $e_y$. We term the $k$-nearest candidates to $e_x$ except $e_y$ ``opponent entities'', as they are also possible aligned entities to $e_x$. These entities can provide additional information for refusing or accepting the input entity pair $[e_x, e_y]$.

\paragraph{Action}
An action $a$ is a binary number that represents the agent's choice. The binary schema has two advantages. First, the corresponding best policy can be an easier function to be approximated. Selecting one entity from multiple candidates is significantly more complex than judging a pair of entities, especially for the case of existing thousands of different candidates. On the other hand, the binary schema enables the agent to suspend the current candidate pair. For example, if the agent chooses \emph{mismatch}, the source entity still has a chance to be correctly matched in the following interactions. By contrast, in the classification schema, the agent must select one entity as the final choice.

\paragraph{Policy}
The policy $\pi(a|s, \theta)$ is parameterized by graph neural networks (GNNs), where $\theta$ denotes the parameter set. We illustrate its architecture in Figure \ref{fig:agent}. Given a state $s = [e_x, e_y]$, we first extract the features by a multi-layer GNN. Here, we use vanilla GCN~\cite{GCNs} for graph convolution, but other GNN models like GATs~\cite{GAT} can also be employed. The output embedding of $e_x$ at layer $k$ is defined as:
\begin{align}
	\rvg_x^k = \sigma (\sum_{e_i \in N(e_x) \cup \{e_x\}} \frac{1}{c_{x}} \rmW^k_g \rvg_{i}^{k-1})
\end{align}
where $g_x^k$ denotes the output hidden of layer $k$ for $e_x$. $c_x$ is the normalization constant. $\rmW^k_g$ is the weight matrix at layer $k$. $\sigma(\cdot)$ is the activation function (ReLU~\cite{ReLU} in our implementation). For the first layer, we set $\rvg_{i}^{0} = \rve_i$, where $\rve_i$ denotes the input embedding of $e_i$. GCNs efficiently aggregate the neighborhood and self-information into a single vector, which is supposed to be more robust and informative than directly using the trained embeddings. Furthermore, GCNs also allow RLEA to reweight entity embeddings for sequential EEA. For simplicity, we denote the output of the last GCN layer by $\rvg_x$.

Next, we use a linear layer to combine the output embeddings $\rvg_x$, $\rvg_y$, which can be written as follows:
\begin{align}
	\rvh_{e_x, e_y} = \sigma(\rmW_h (\rvg_x || \rvg_y) + \rvb_h),
\end{align}
where $||$ is the concatenation operator to concat $\rvg_x$, $\rvg_y$ to one hidden vector. $\rmW_h$ and $\rvb_h$ are the weight matrix and bias vector, respectively.

We also take the mutual information $I(e_x, e_y)$~\cite{MINE} as an additional feature. Unlike the cosine similarity that weights the difference of two vectors at each dimension, mutual information values more on the high-level correlations. Therefore, it is especially appropriate for the EEA task, where two aligned entities may not have identical neighborhoods due to the heterogeneity. Following~\cite{SSL}, we leverage a neural function $f(\rvg_x, \rvg_y)$ to estimate the density ratio: 
\begin{align}
	 f(\rvg_x, \rvg_y) = \exp(\rvg_x^T \rmW_f \rvg_y),
\end{align}
where $\rmW_f$ is the weight matrix. As aforementioned, we consider opponent entities a kind of future information to aid the agent in making decisions. This idea can be naturally reified by viewing the opponents as negative examples:
\begin{align}
	\hat{I}_{e_x, e_y} = \frac{f(\rvg_x, \rvg_y)}{\sum_{e_i \in O_y \cup \{e_y\}} f(\rvg_x, \rvg_i)}.
\end{align}
The above equation has a similar form to that used in InfoNCE~\cite{SSL}. But from another aspect, $\hat{I}_{e_x, e_y}$ can be also understood as the probability of outputting the action \emph{match} based on the mutual information estimator (MIE).

Finally, we concatenate all estimates to obtain the final action distribution:
\begin{align}
	\rvp_{e_x, e_y} &= \textit{Softmax} (\rmW_p (\rvh_{e_i, e_j}||\hat{I}_{e_x, e_y})), \nonumber\\
					&= \pi(a|s, \theta)
\end{align}
where $\rvp_{e_x, e_y}$ is the normalized action distribution . $\rmW_p$ is the weight matrix.

\paragraph{Reward}
We assign the reward for the given output action $a$ by the following equation:
\begin{align}
	r = \begin{cases}
		1,\qquad &\text{a \emph{true match}},\\
		-10,\qquad &\text{a \emph{false mismatch}},\\
		0,\qquad &\text{elsewise}.
	\end{cases}
\end{align}  
The goal of the agent is to maximize the overall reward, i.e., output $1$ as much as possible for aligned pairs. Therefore, we set a positive reward for $a=l$ when the input pair match and a severe penalty ($-10$ is most efficient in our implementation) for a \emph{false mismatch}. For other cases, the agent will receive a reward $0$, as they do not directly increase or decrease the number of alignment pairs.

\paragraph{Optimization}
We use the policy gradient algorithm REINFORCE~\cite{REINFORCE} to find the parameters leading to a larger reward. To reduce the variance, we employ a baseline function for comparison. Therefore, the gradient at $i$-step in an episode is:
\begin{align}
	\label{eq:reinforce}
	\nabla \theta = \alpha \gamma^i \delta \nabla \ln \pi(a|[e_x, e_y]),
\end{align}
where $\alpha$ is the learning step-size. $\gamma$ is the discount factor. $\delta$ is the relative advantage of policy $\pi(a|[e_x, e_y])$ than the baseline, i.e., how much better the output action $a$ is than mean or random. It can be defined as follows:
\begin{align}
	\delta &= G - \hat{v}([e_x, e_y]) \nonumber \\ 
	       &= \sum_{k=i+1}^{T} \gamma^{k-i-1} r_k - \hat{v}([e_x, e_y]),
\end{align}
where $G$ is the return based on the future rewards. $T$ denotes the episode length. The baseline function $\hat{v}([e_x, e_y])$ in this paper is an estimate of the state value.

\subsection{Environment}
Generally, the environment for an RL task should conform with three basic properties: dependency, dynamics, and difficulty.

\paragraph{Dependency}
The output action may change the later states. For sequential entity alignment, a \emph{true match} will not only yield a correct alignment, but also exclude some plausible candidates for the following judgments, contributing to higher overall reward. Even a \emph{false match} also has its value in filtering out two wrong entities. Therefore, we should consider the long-term dependencies. 

To this end, for each entity $e_x$ in $\gG_1$, its $k$-nearest entities $e_1, e_2, ..., e_k$ in $\gG_2$ are selected as candidates. Those entities are then concatenated with $e_x$ to form $k$ candidate pairs $[e_x, e_1], [e_x, e_2], ..., [e_x, e_k]$. The environment maintains a sequence $c_1, c_2, ..., c_j$, in which each element is such a candidate pair. At the $i$-th step, the environment pops a candidate pair $[e_x, e_y]$ as $s_i$. If it receives an action $a_i=1$ from the agent, all candidate pairs containing $e_x$ or $e_y$ will be removed from the sequence.

\paragraph{Dynamics}
The environment is usually dynamic. The state-action sequences are different in different episodes. A dynamic environment makes the agent capable of capturing the general rules of the game, which is crucial to avoid overfitting. For the EEA task, if the state sequence is constant at each training episode, the agent will fit this sequence. However, the states are entirely different at the testing phase. 

To ensure the dynamic property, we set a skip rate $p_s$. The environment randomly skips a candidate pair with probability $p_s$ and then pops the next pair. Therefore, the length and elements of the state sequence change in each episode.

\paragraph{Difficulty}
Often, the difficulty of a game is improved gradually as step number grows. For example, the health and speed of enemies in video games usually increase over game time. On the other hand, it is also a general strategy to break down complex knowledge by a sequence of learning episodes of increasing difficulty, which is known as curriculum learning~\cite{CurriculumLearning}.

For sequential entity alignment, the difficulty of a candidate pair can be estimated based on the cosine similarity of the two entities and their label, which can be written as follows:
\begin{align}
	\label{eq:difficulty}
	d({e_x, e_y}) =&\,l(C_{e_x, e_{max}} -  C_{e_x, e_y}) \nonumber \\
	&+ (1-l)(\tau - C_{e_x, e_{max}} + C_{e_x, e_y})
\end{align}
where $C_{e_x, e_y}$ is the cosine similarity between $e_x$ and $e_y$. $e_{max}$ denotes the entity with the largest similarity to $e_x$. We use the difference between $C_{e_x, e_{max}}$ and $C_{e_x, e_y}$ as the basis to estimate the extent, and the label $l$ as the sign. When $l=1$, i.e., the first term in Eq. (\ref{eq:difficulty}), a large difference between $C_{e_x, e_y}$ and $C_{e_x, e_{max}}$ will result in high difficulty because $e_x$, $e_y$ may be too dissimilar with each other. The situation is reversed for $l=0$. We add a hyper-parameter $\tau$ to balance difficulty scores between these two cases.

Then, we can sort candidate pairs by the difficulty in ascending order, such that the agent will always start from the relatively easier states. However, this operation is inapplicable to the testing set where the label information is unknown. To mitigate this problem, we propose a curriculum learning strategy. We do not directly sort the candidate pairs by difficulty score. Instead, we sort them based on the cosine similarity and re-weight the skip rate $p_s$ for each pair by its normalized difficulty score. Therefore, for each episode, the agent will start from pairs with high similarity, and the more difficult states will be skipped with larger probabilities. As the policy is optimized, we gradually decrease $p_s$ to approximate the testing environment. The final state sequence shall have a similar arrangement to that at the testing phase. 

The skip rate $p_s^{i,t}$ at the $i$-th step in episode $t$ can be written as:
\begin{align}
	\label{eq:final_skip_rate}
	p_s^{i,t} = max(p_s^{min}, \eta^{t-1} p_s d_i ),
\end{align}
where $p_s^{min}$ is the minimal skip rate to ensure the dynamic property. $p_s$ is the basic skip rate. $d_i$ denotes the difficulty of state $s_i$ at $i$-th step. As episode number grows, $p_s^{i,t}$ decreases with discount factor $\eta$ exponentially until it meets the lower bound $p_s^{min}$.

\begin{figure}[t]
	\centering
	\includegraphics[width=.95\linewidth]{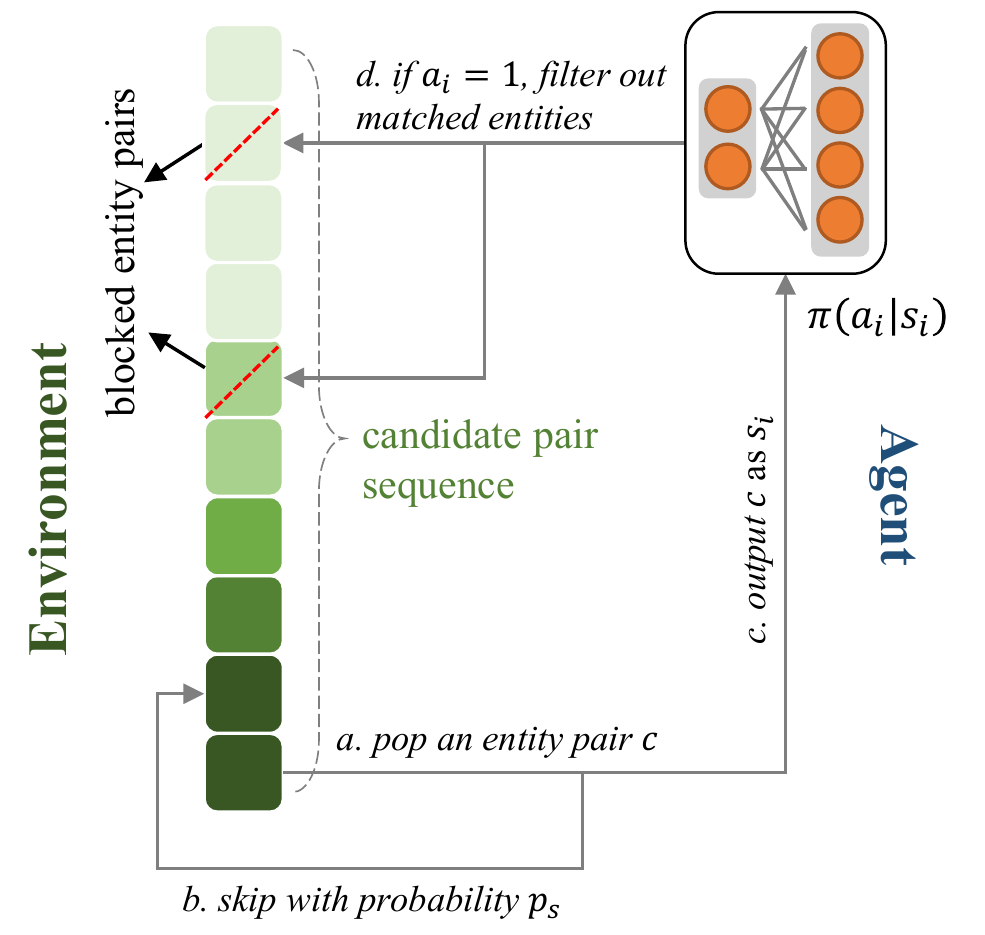}
	\caption{Illustration of how the environment interacts with the agent in RLEA.}
	\label{fig:arch}
\end{figure}
We illustrate how the environment collaborates with the agent in Figure~\ref{fig:arch}: a. the environment pops an entity pair $c$ from the candidate pair sequence; b. this entity pair may be skipped with probability $p_s$ (Equation (\ref{eq:final_skip_rate})); c. the non-skipped pair is outputted by the environment as $s_i$; d. the agent takes $s_i$ as input, and its output action changes the candidate pair sequence reversely. The detailed implementation can be found in Appendix~\ref{app:impl}.

\begin{table*}[t]
	\centering
	\caption{Hits@1 results on four datasets (5-fold cross-validation).}
	\label{tab:main_results}
	\resizebox{.9\linewidth}{!}{
		\begin{tabular}{lcccccccccccc}
			\toprule
			\multirow{2}{*}{Methods} & \multicolumn{3}{c}{EN-FR} & \multicolumn{3}{c}{EN-DE} & \multicolumn{3}{c}{D-W} & \multicolumn{3}{c}{D-Y}\\
			\cmidrule(lr){2-4} \cmidrule(lr){5-7} \cmidrule(lr){8-10} \cmidrule(lr){11-13} 
			& Orig & Seq & RLEA & Orig & Seq & RLEA & Orig & Seq & RLEA & Orig & Seq & RLEA \\
			\midrule
			JAPE~\cite{JAPE} & .247 & .291 & \textbf{.322} & .307 & .332 & \textbf{.336} & .259 & .279 & \textbf{.301} & .463 & .547 & \textbf{.607} \\
			SEA~\cite{SEA} & .280 & .317 & \textbf{.365} & .530 & .556 & \textbf{.571} & .360 & .359 & \textbf{.414} & .500 & .564  & \textbf{.643} \\
			RSN~\cite{RSN} & .393 & .410 & \textbf{.429} & .587 & .614 & \textbf{.634} & .441 & .466 & \textbf{.493} & .514 & .546 & \textbf{.566} \\
			RDGCN~\cite{RDGCN} & .755 & .801 & \textbf{.830} & .830 & .861 & \textbf{.878} & .515 & .517 & \textbf{.541} & .931 & .951 & \textbf{.974} \\
			\bottomrule
		\end{tabular}
	}
\end{table*}
\section{Experiment}
\label{sec:experiment}

We conducted experiments to verify the effectiveness of the proposed RL-based framework. The trained entity embeddings were obtained from the OpenEA project \footnote{https://github.com/nju-websoft/OpenEA}.

\subsection{Dataset Settings}

We used the 15K benchmark proposed by OpenEA. It consists of four subsets: EN-FR, EN-DE, D-W, and D-Y. The former two are cross-lingual datasets, where EN, FR, DE denote English, French, and German versions of DBpedia, respectively. The latter two are cross-source datasets, where D, W, Y denote DBpedia~\cite{DBpedia}, WikiData~\cite{Wikidata}, and Yago~\cite{Yago}, respectively. We used ``V1'' subsets that has similar distributions to original KGs. Please refer to~\cite{OpenEA} for detailed statistics.

\subsection{Compared Methods}
We select the following methods as baselines:
\begin{itemize}
	\item JAPE~\cite{JAPE}, which learns attribute embeddings and relational embeddings jointly for EEA.
	\item SEA~\cite{SEA}, which adopts adversarial learning to learn the projection matrix.
	\item RSN~\cite{RSN}, which leverages recurrent neural networks (RNNs)~\cite{RNN} to learn KG embeddings.
	\item RDGCN~\cite{RDGCN}, which uses GCNs to capture the neighborhood information into entity embeddings.  
\end{itemize}

We also design a basic sequential strategy called \emph{Seq} for comparison. We follow the algorithm used in BootEA~\cite{BootEA} to implement it. Entity pairs with similarity above a predefined threshold are regarded as \emph{match}, or the algorithm randomly chooses actions based on cosine similarity.

\subsection{Main Results}

The results on four datasets are shown in Table~\ref{tab:main_results}. Orig denotes the original results of the EEA methods. We can observe that RLEA significantly improved the performance of all baseline methods, including the best-performing one, RDGCN. Therefore, we believe that RLEA provides a better way to exploit aligned entity pairs from embeddings than the widely-used heuristic strategy.

Specifically, the performance improvement of JAPE is most notable, with $30.4\%$ and $31.1\%$ increases on EN-FR and D-Y, respectively. Although RSN has minimal performance increase, the difference is still significant. Similarly, RDGCN also achieved better performance, leading to a new state-of-the-art on the benchmark.

With the basic sequential strategy Seq, four baseline methods also achieved better Hits@1 results on most datasets except D-W. This observation empirically proves the advantages of modeling EA as a sequential decision-making task.

Note that RLEA also has its limitations. We find there exists a dataset bias. The performance improvement on EN-FR dataset is notable, but that on D-W is less significant. We believe that the sequential evaluation process might cause this bias. For instance, Seq also got worse performance than the original method SEA on D-W. We leave how to mitigate this problem in future work.

\begin{figure}[t]
	\centering
	\includegraphics[width=\linewidth]{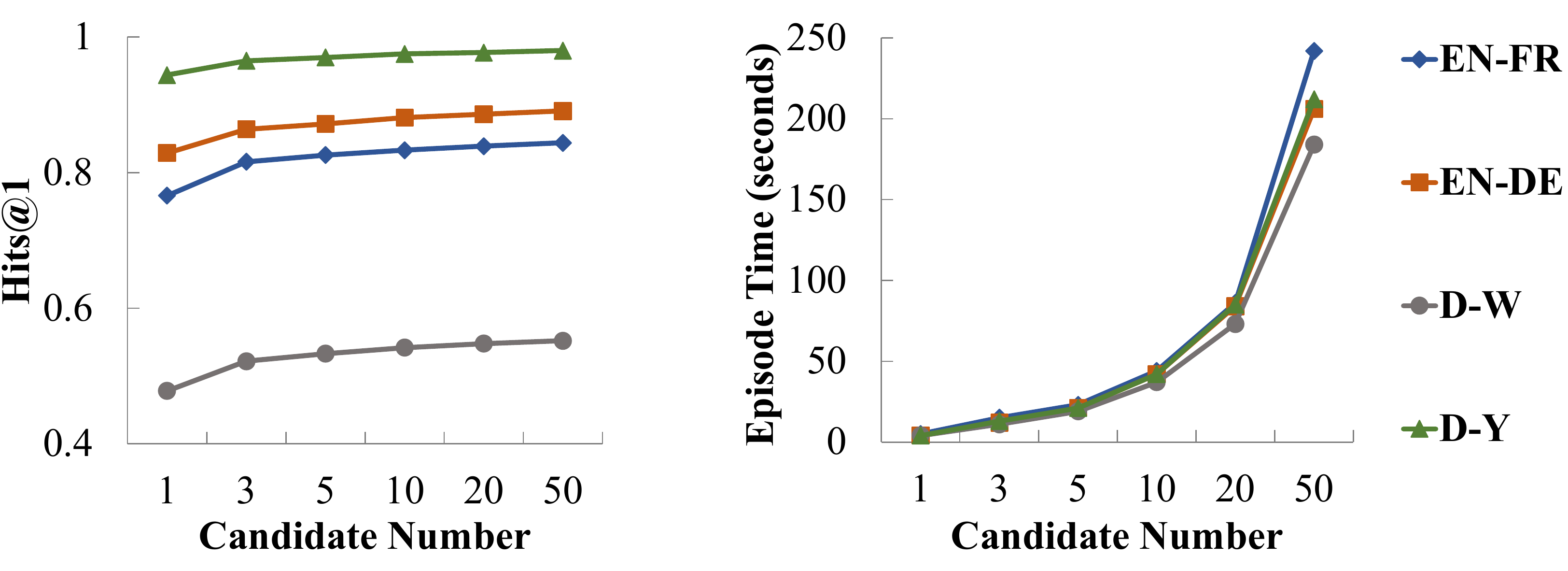}
	\caption{Hits@1 and episode time w.r.t. candidate number on four datasets.}
	\label{fig:candidate_number}
	\vspace{-1.2em}
\end{figure}

\subsection{Influence of Candidate Number}

In RLEA, the candidate number for each entity is an important hyper-parameter as it decides the length of the candidate sequence. A large value means covering more correct alignment pairs as well as more plausible pairs. Therefore, it is necessary to study how this hyper-parameter influences the performance of RLEA. 

We used the embeddings of the best-performing method RDGCN as input in this experiment. As shown in the left of Figure~\ref{fig:candidate_number}, the Hits@1 results on four datasets gradually increase with candidate number from 1 to 5, but converge after 10. When candidate number was set to 1, for each entity, only the pair with the highest similarity was added to the sequence, resulting in similar or even worse results compared with the original method. For example, on D-W, the hits@1 of RLEA is $0.478$, significantly below that of the original RDGCN ($0.541$). As the candidate number increased, more aligned pairs were added to the sequence, the performance improved steadily. It then gets saturated due to more unaligned pairs were also added to the sequence. 
\begin{figure}[t]
	\centering
	\includegraphics[width=\linewidth]{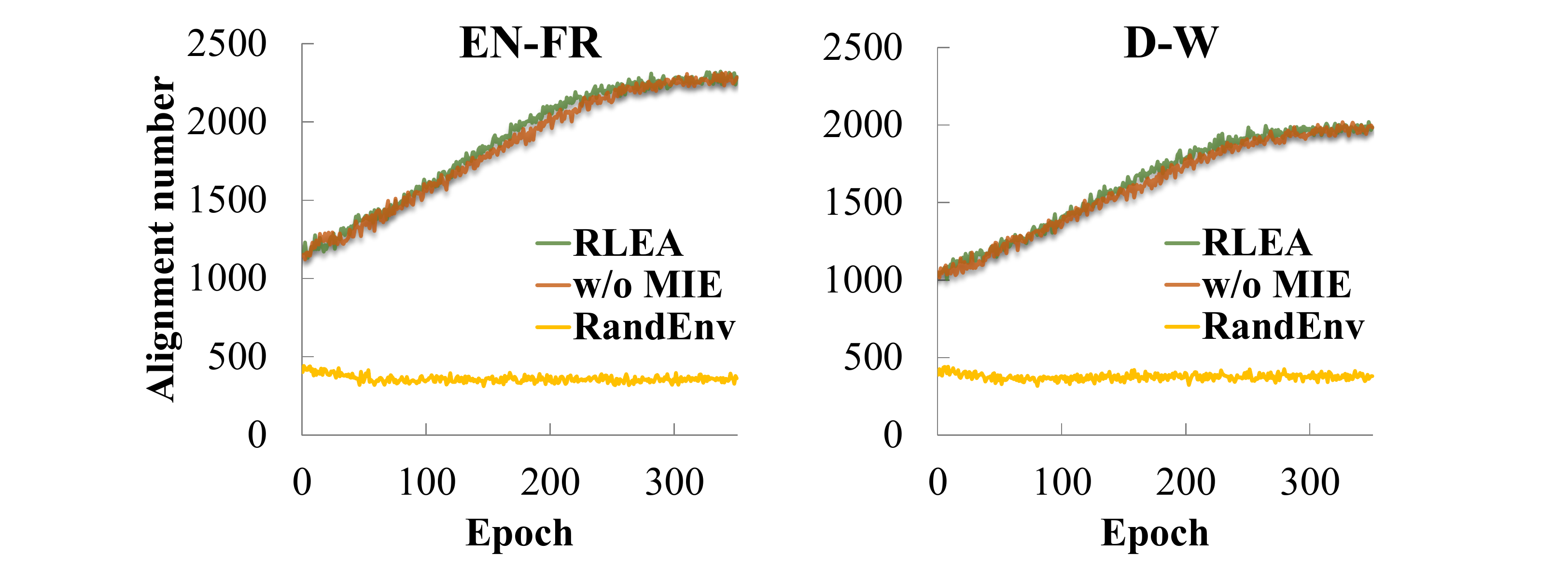}
	\caption{Alignment number w.r.t. episode number, on EN-FR and D-W datasets.}
	\label{fig:env}
	\vspace{-1.2em}
\end{figure}
On the right of Figure~\ref{fig:candidate_number}, we show the runtime of one testing episode w.r.t. candidate number, which, however, grows exponentially. This observation suggests that setting a large candidate number is computationally expensive. Therefore, we decide to use the top-$10$ candidates in our implementation, for sake of performance and efficiency.

\begin{table*}[t]
	\centering
	\caption{Comparing RLEA with conventional methods.}
	\label{tab:conventional_results}
	\resizebox{0.9\linewidth}{!}{
		\begin{tabular}{lcccccccccccc}
			\toprule
			\multirow{2}{*}{Methods} & \multicolumn{3}{c}{EN-FR} & \multicolumn{3}{c}{EN-DE} & \multicolumn{3}{c}{D-W} & \multicolumn{3}{c}{D-Y}\\
			\cmidrule(lr){2-4} \cmidrule(lr){5-7} \cmidrule(lr){8-10} \cmidrule(lr){11-13} 
			& Precision & Recall & F1-score & Precision & Recall & F1-score & Precision & Recall & F1-score & Precision & Recall & F1-score \\
			\midrule
			LogMap  & .818 & .729 & .771 & .925 & .725 & .813 & - & - & - & .960 & .943 & .951 \\
			PARIS  & \textbf{.907} & \textbf{.900} & \textbf{.903} & \textbf{.938} & \textbf{.933} & \textbf{.935} & \textbf{.746} & \textbf{.723} & \textbf{.734} & .875 & .868  & .872 \\ \midrule
			OpenEA  & .755 & .755 & .755 & .830 & .830 & .830 & .572 & .572 & .572 & .931 & .931 & .931 \\
			RLEA & .830 & .830 & .830 & .878 & .878 & .878 & .611 & .611 & .611 & \textbf{.974} & \textbf{.974} & \textbf{.974} \\
			\bottomrule
		\end{tabular}
	}
	
\end{table*}

\begin{figure*}[thb]
	\centering
	\includegraphics[width=\linewidth]{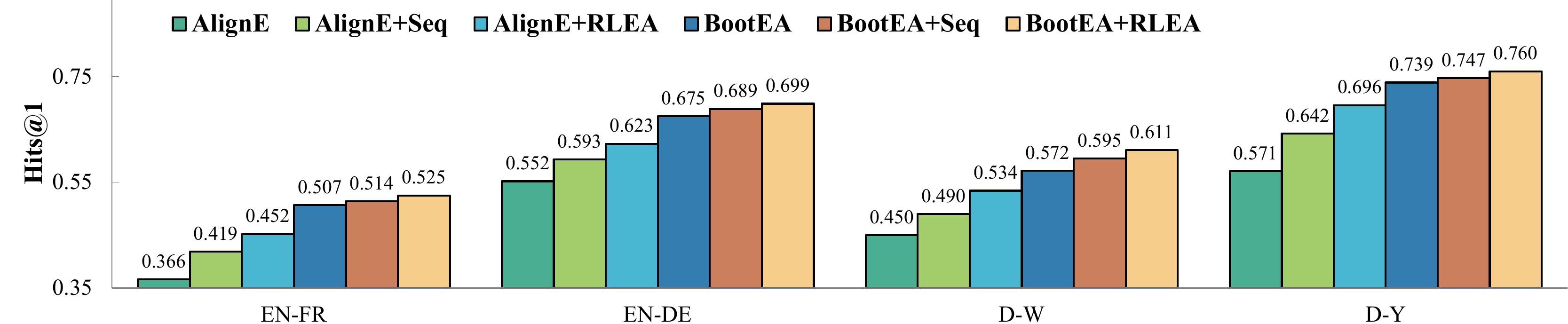}
	\caption{A comparison between RLEA and BootEA.}
	\label{fig:bootea_rlea}
	\vspace{-1.2em}
\end{figure*}

\subsection{Effectiveness of Modules}

We conducted experiments to verify the effectiveness of mutual information estimator (MIE) and the proposed environment. We developed two variants of RLEA: (1) RLEA without MIE (denoted as w/o MIE), and (2) RLEA with a random environment (RandEnv). The random environment still maintains a candidate pair sequence but does not have the difficulty and skipping settings. All candidate pairs in the sequence are randomly reset at the start of each episode.

We compare the results in Figure \ref{fig:env}, from which we find that the agent does not work in the random environment on all datasets. The alignment number even slowly decreases during training. This is because that the state sequence in the random environment changes irregularly. The agent fails to establish an effective policy to maximize the reward for all episodes. Furthermore, the random environment does not have a curriculum learning strategy to help the agent study from easy to hard. Therefore, the agent is not able to capture the general rules in the random environment.

On the other hand, we find that MIE slightly improves the performance. It does not have a significant advantage over the final reward or alignment number. This may be because that the output embeddings of GNNs have already included sufficient information to judge entity pairs. Nevertheless, the estimation provided by MIE helps the agent find the best policy rapidly, which is crucial when applying to larger datasets. A more detailed version of Figure~\ref{fig:env} is shown in Appendix~\ref{app:ablation_study}, from which we can obtain the consistent observations.

\subsection{A Comparison of RLEA and BootEA}
\label{app:bootea}

BootEA~\cite{BootEA} is a bootstrapping method that iteratively labels possible entity alignment as training data. Like RLEA, BootEA assumes a candidate entity should not be aligned twice. Therefore, it is interesting to compare and discuss these two methods. 

We illustrate the experimental results on four datasets in Figure~\ref{fig:bootea_rlea}. AlignE is a variant of BootEA without bootstrapping process. We first compare the left four columns. Obviously, BootEA (4th column) has a better effect in improving the performance of AlignE, as it directly participates in training AlignE by iteratively adding plausible alignment pairs. In contrast, Seq (2nd column) and RLEA (3rd column) only use the trained embeddings as input and do not modify the embeddings or training procedure. They are thus more extensible and applicable to arbitrary EEA methods. 

In fact, it is no contradiction to integrate these two types of methods. The performance improvement (5th and 6th columns) is still significant and consistent on all four datasets.

\subsection{Competing with Conventional Methods}
\vspace{-0.5em}

There has always been an argument about the practical use of EEA. Most EEA methods are end-to-end and easy to be deployed. The performance also improves when new models are developed. However, a significant performance gap still exists between EEA methods and those conventional methods like Paris~\cite{PARIS} and LogMap~\cite{LogMap}. We show in Table~\ref{tab:conventional_results} that RLEA with the embeddings of best EEA methods as input can narrow this gap and even outperform the conventional methods on some datasets.
As shown in Table~\ref{tab:conventional_results}, PARIS is the best method that outperformed others on all datasets except D-Y. However, The second method changed from LogMap to RLEA. We can find that RLEA not only outperformed LogMap on EN-FR, but also achieved the best performance on D-Y. 

We should notice that the alignment pairs exploited by EEA methods and conventional methods are not all overlapped~\cite{OpenEA}. It is possible to integrate them to achieve better performance~\cite{JAPE,OpenEA}. In this sense, RLEA is also the best choice to be combined with conventional methods.

\section{Conclusion and Future Work}
\label{sec:conclusion}
\vspace{-0.5em}
In this paper, we proposed an RL-based entity alignment framework, which can advance most existing EEA methods without modifying their parameter settings or infrastructures. Our experiments demonstrate consistent and significant improvement on all baseline methods. We plan to study how to jointly train EEA methods and RLEA for further improvement in future work.

\section*{Acknowledgement}
\label{sec:ack}
\vspace{-0.5em}
We want to thank the anonymous reviewers for their invaluable comments. This work is funded by NSFCU19B2027/NSFC91846204, National Key R\&D Program of China (Funding No.SQ2018YFC000004), Zhejiang Provincial Natural Science Foundation of China (No.LGG22F030011).

\bibliography{rlea}
\bibliographystyle{acl_natbib}

\clearpage
\appendix

\section{Implementation}
\label{app:impl}
\subsection{Algorithm}

We show the training procedure of RLEA by Algorithm~\ref{alg:rlea}. The input is two KGs, trained embeddings of an arbitrary EEA method, and parameter settings. If the EEA method has projection matrices~\cite{MTransE, SEA}, the embeddings of $\gG_2$ should be projected to the space of $\gG_1$ by the corresponding matrix before the training starts. We first initialize all parameters of the policy network. The episode sequence and candidate pair sequence will be reset at the start of each episode. After that, the agent interacts with the environment, which generates a state-action sequence. We then use REINFORCE algorithm to update the policy network with the generated sequence.
\begin{algorithm}[th]
	\caption{RLEA}
	\label{alg:rlea}
	\LinesNumbered
	\KwIn{Two KGs $\gG_1$, $\gG_2$, entity embeddings $\rmE_1$, $\rmE_2$, and number of episodes $N$.}
	\KwOut{The policy $\pi(a|s, \theta)$.}
	
	Initialize the policy parameter $\theta$;
	
	\For{t=1:N}{
		Reset episode sequence to empty and initialize candidate pair sequence\;
		
		\Repeat{All entities are matched or candidate pair sequence is empty.}{
			Pop a candidate pair $c$ from candidate pair sequence\;
			Calculate the skip rate $p_s$ with Equation~(\ref{eq:final_skip_rate})\;
			
			$\mu \leftarrow \text{RandInt}(0, 1, p_s)$\;

			\lIf{$\mu = 1$}{
				\textbf{continue}
			}
			\lElse{
				$s \leftarrow c$
			}
			$a \sim \pi(a|s, \theta)$\;
			
			\lIf{$a=1$}{
				Update candidate pair sequence according to $c$
			}
			Add $s,a$ to episode sequence\;
		}
		\For{each step in the episode sequence}{
			Update the policy parameter $\theta$ according to Equation~(\ref{eq:reinforce}).
		}
	}
	Output $\pi(a|s, \theta)$.
\end{algorithm}

\subsection{Parameter Settings}
For each EEA method, we directly used their trained entity embeddings as input and did not modify these vectors during training. The embedding-size was identical to that used in OpenEA~\cite{OpenEA}. The number of training episodes was set to 500, and the learning step-size was set to 0.0001. The candidate pair number for each entity was set to 10. 

\section{Detailed Results of Ablation Study}
\label{app:ablation_study}

\begin{table*}[t]
	\centering
	\caption{Hits@1 results on OpenEA 100K datasets.}
	\label{tab:100k_results}
	\resizebox{.9\linewidth}{!}{
		\begin{tabular}{lcccccccccccc}
			\toprule
			\multirow{2}{*}{Methods} & \multicolumn{3}{c}{EN-FR} & \multicolumn{3}{c}{EN-DE} & \multicolumn{3}{c}{D-W} & \multicolumn{3}{c}{D-Y}\\
			\cmidrule(lr){2-4} \cmidrule(lr){5-7} \cmidrule(lr){8-10} \cmidrule(lr){11-13} 
			& Orig & Seq & RLEA & Orig & Seq & RLEA & Orig & Seq & RLEA & Orig & Seq & RLEA \\
			\midrule
			JAPE~\cite{JAPE} & .165 & .172 & \textbf{.197} & .152 & .162 & \textbf{.169} & .211 & .229 & \textbf{.257} & .287 & .308 & \textbf{.323} \\
			SEA~\cite{SEA} & .225 & .229 & \textbf{.261} & .341 & .345 & \textbf{.376} & .291 & .293 & \textbf{.338} & .490 & .525  & \textbf{.545} \\
			\bottomrule
		\end{tabular}
	}
\end{table*}
 \begin{figure*}[ht]
	\centering
	\includegraphics[width=.9\linewidth]{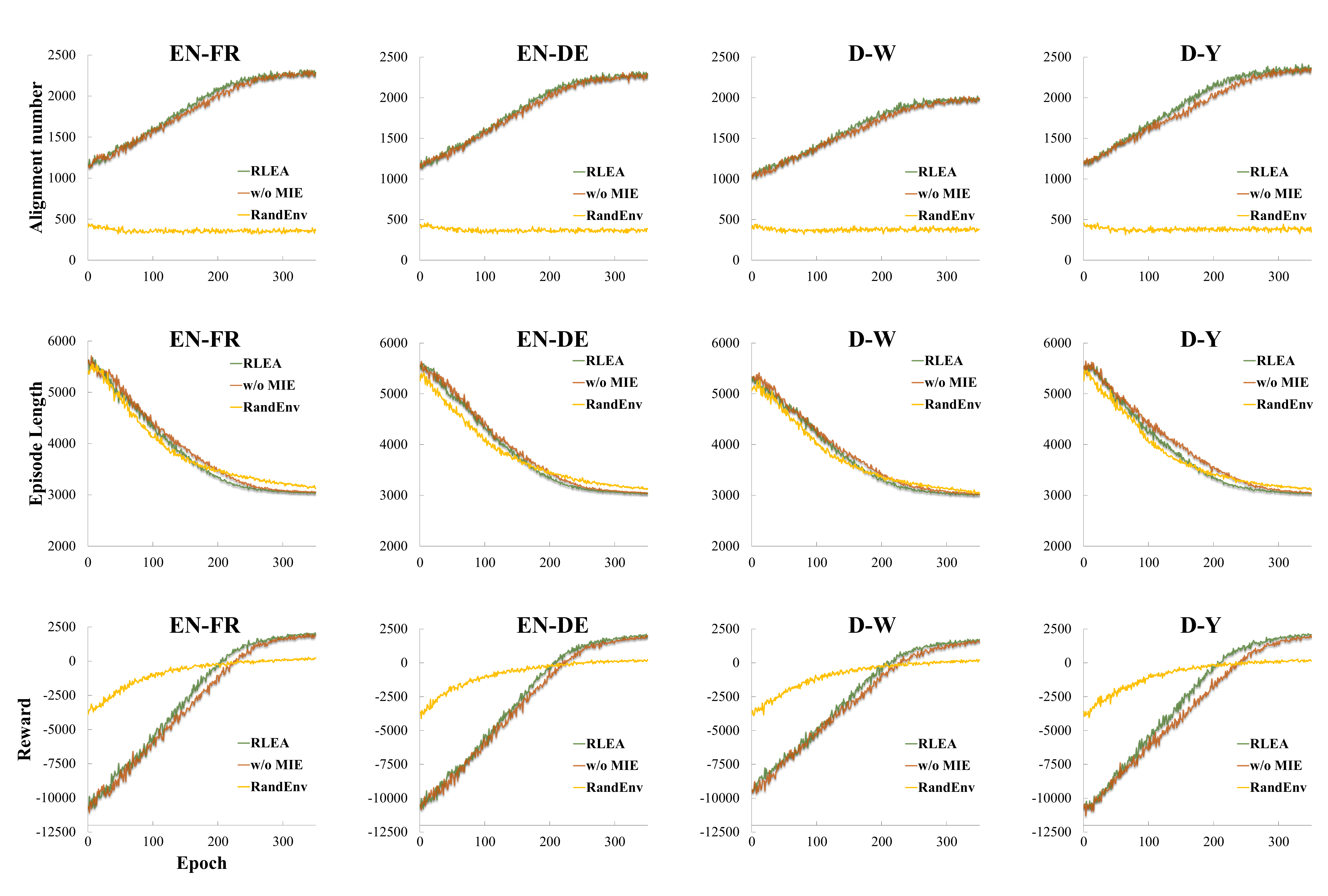}
	\caption{Alignment number, episode length, and reward w.r.t. episode number, on four datasets.}
	\label{fig:full_ablation_study}
\end{figure*}

The detailed results of RLEA and its two variants are shown in Figure~\ref{fig:full_ablation_study}. Overall, the full RLEA still has the best performance and training speed, especially on D-Y dataset. The method without MIE also have competitive performance on four datasets, which demonstrates the effectiveness of the RL-based sequential EEA.

From the bottom sub-figures, we find that the agent tries to find a policy to achieve high rewards in the random environment. However, $0$ is almost the best reward it can get. The agent fails to establish a good policy in this dynamic environment. 

\section{Results on OpenEA 100K datasets}
As shown in Table~\ref{tab:100k_results}, the Hits@1 results on OpenEA 100K datasets are consistent with those on 15K datasets. RLEA still outperformed the baselines on four datasets. We did not consider RDGCN and RSN in this experiment, as they can only be trained on CPUs (confirmed from the authors of OpenEA).

\end{document}